\documentclass{article}

\usepackage{times}
\usepackage{fancyvrb}

\title{An Alternative to RDF-Based Languages for the Representation and Processing of Ontologies in the Semantic Web}

\author{\large J Gerard Wolff \\
	\normalsize CognitionResearch.org.uk \\
	\normalsize Wales, UK \date{}}

\begin{document}

\maketitle

\begin{abstract}

\noindent This paper describes an approach to the representation and processing of ontologies in the Semantic Web, based on the ICMAUS theory of computation and AI. This approach has strengths that complement those of languages based on the Resource Description Framework (RDF) such as RDF Schema and DAML+OIL. The main benefits of the ICMAUS approach are simplicity and comprehensibility in the representation of ontologies, an ability to cope with errors and uncertainties in knowledge, and a versatile reasoning system with capabilities in the kinds of probabilistic reasoning that seem to be required in the Semantic Web.

\end{abstract}

\section{Introduction}

One of the most popular approaches to the representation and processing of ontological knowledge in the Semantic Web is a family of languages based on the Resource Description Framework (RDF), such as RDF Schema (RDFS), the Ontology Inference Layer (OIL) and DAML+OIL (see, for example, \cite{broekstra_etal_2002,mcguinness_2002,horrocks_2002,fensel_etal_2001}). For the sake of brevity, these languages will be referred to as `RDF+'.

This paper describes an alternative approach to the representation and processing of ontologies based on the ICMAUS\footnote{Short for {\em information compression by multiple alignment, unification and search}} theory of computation and AI. An overview of the theory is presented in \cite{wolff_icmaus_overview,wolff_2001_igpl} with more detail in \cite{wolff_2000,wolff_1999_comp,wolff_1999_prob}.

In this context, the word `alternative' is intended as an inclusive OR, not an exclusive OR. The RDF+ and ICMAUS approaches appear to have complementary strengths and may be used together. In the spirit of evolvability in the Web, they may be developed in parallel.

The main benefits of the ICMAUS approach appear to be:

\begin{itemize}

\item Simplicity and comprehensibility in the way knowledge is represented.

\item An ability to cope with errors and uncertainties in knowledge.

\item A versatile reasoning system with capabilities in the kinds of probabilistic reasoning that seem to be required in the Semantic Web. 

\end{itemize}

In what follows, the ICMAUS theory is first described in outline. Then an example is presented showing how an ontology may be represented and processed in the ICMAUS framework. This is followed by a comparison with RDF+.

\section{The ICMAUS theory}

The ICMAUS theory grew out of a long tradition in psychology that many aspects of perception, cognition and the workings of brains and nervous systems may be understood as information compression. It is founded on principles of Minimum Length Encoding (MLE), pioneered by \cite{solomonoff_1964,wallace_boulton_1968,rissanen_1978} and others (see \cite{li_vitanyi_1997}).

The theory is conceived as an abstract model of {\em any} system for processing information, either natural or artificial. In broad terms, it receives data (designated `New') from its environment and adds these data to a body of stored of knowledge (called `Old'). At the same time, it tries to compress the information as much as possible by searching for full or partial matches between patterns and unifying patterns or sub-patterns that are the same. In the process of compressing information, the system builds `multiple alignments' of the kind shown below.

The ICMAUS framework is partially realised in the SP61 computer model that builds multiple alignments but requires its store of Old knowledge to be supplied by the user. It also calculates probabilities for inferences that can be drawn from the multiple alignments. The SP70 model, which is less mature, is an augmented version of SP61 that embodies all elements of the framework including a process of learning by building the repository of Old, stored knowledge. Both models have polynomial time complexity.

The ICMAUS framework is Turing-equivalent in the sense that it can model a universal Turing machine \cite{wolff_1999_comp} but it can perform a range of AI-style operations without the kind of programming that would be needed to make a Turing machine perform in the same way.

To date, the main areas to which the ICMAUS framework has been applied are probabilistic reasoning \cite{wolff_1999_prob}, parsing and production of natural language \cite{wolff_2000}, fuzzy pattern recognition and best-match information retrieval \cite{wolff_1999_prob}, planning and problem solving, modelling concepts in logic and mathematics, and unsupervised learning \cite{wolff_icmaus_overview,wolff_2001_igpl}.

\section{The representation and processing of ontologies in the ICMAUS framework}\label{representation_and_processing}

In the ICMAUS framework, {\em all} knowledge is represented by arrays of symbols in one or two dimensions termed {\em patterns}. In work to date, the main focus has been on one-dimensional patterns. A {\em symbol} in this context is a `mark' that can be matched in a yes/no manner with other symbols but has no intrinsic meaning. In the example below, each symbol is represented by a string of non-space characters.

As in bioinformatics, a multiple alignment in the ICMAUS framework is an arrangement of two or more sequences of symbols in rows (or columns) so that, by judicious `stretching' of sequences where necessary, symbols that match each other are arranged in columns (or rows). By contrast with the bioinformatics concept (where all sequences have the same status), one of the sequences in each alignment is New and the rest are Old. A `good' alignment is one where the New pattern can be encoded economically in terms of the Old patterns, in accordance with MLE principles (see \cite{wolff_2000}).

\begin{figure*}[!htb]
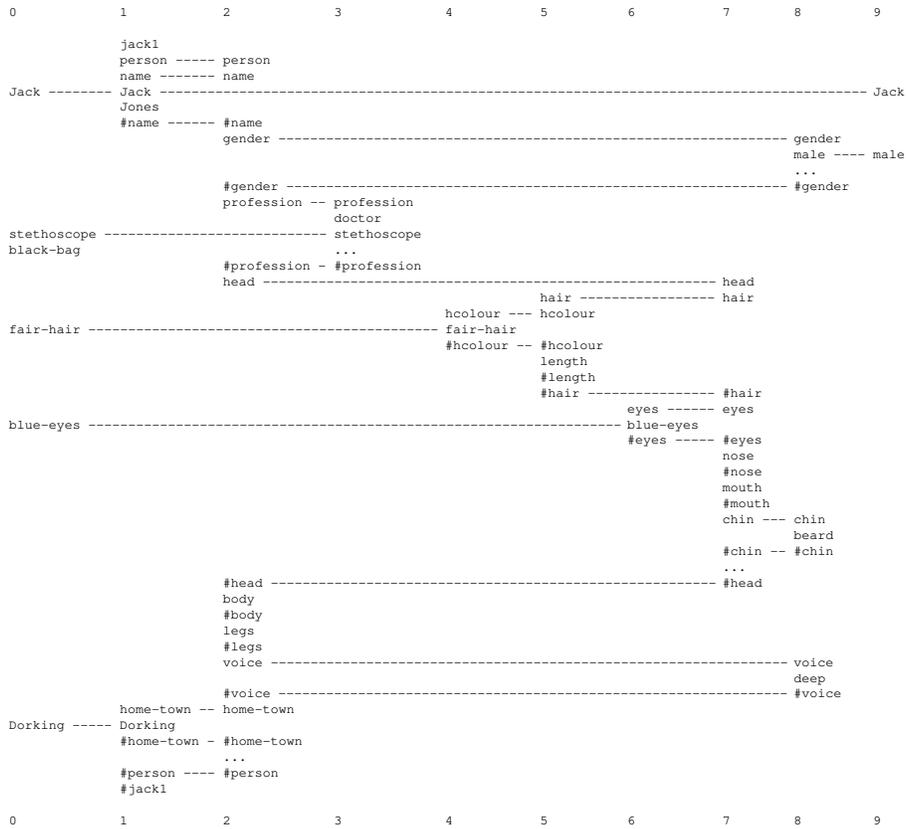

\fontsize{05.00pt}{06.00pt}
\begin{center}
\begin{BVerbatim}
0             1            2             3             4           5          6           7        8         9   

              jack1                                                                                              
              person ----- person                                                                                
              name ------- name                                                                                  
Jack -------- Jack ----------------------------------------------------------------------------------------- Jack
              Jones                                                                                              
              #name ------ #name                                                                                 
                           gender ---------------------------------------------------------------- gender        
                                                                                                   male ---- male
                                                                                                   ...           
                           #gender --------------------------------------------------------------- #gender       
                           profession -- profession                                                              
                                         doctor                                                                  
stethoscope ---------------------------- stethoscope                                                             
black-bag                                ...                                                                     
                           #profession - #profession                                                             
                           head --------------------------------------------------------- head                   
                                                                   hair ----------------- hair                   
                                                       hcolour --- hcolour                                       
fair-hair -------------------------------------------- fair-hair                                                 
                                                       #hcolour -- #hcolour                                      
                                                                   length                                        
                                                                   #length                                       
                                                                   #hair ---------------- #hair                  
                                                                              eyes ------ eyes                   
blue-eyes ------------------------------------------------------------------- blue-eyes                          
                                                                              #eyes ----- #eyes                  
                                                                                          nose                   
                                                                                          #nose                  
                                                                                          mouth                  
                                                                                          #mouth                 
                                                                                          chin --- chin          
                                                                                                   beard         
                                                                                          #chin -- #chin         
                                                                                          ...                    
                           #head -------------------------------------------------------- #head                  
                           body                                                                                  
                           #body                                                                                 
                           legs                                                                                  
                           #legs                                                                                 
                           voice ----------------------------------------------------------------- voice         
                                                                                                   deep          
                           #voice ---------------------------------------------------------------- #voice        
              home-town -- home-town                                                                             
Dorking ----- Dorking                                                                                            
              #home-town - #home-town                                                                            
                           ...                                                                                   
              #person ---- #person                                                                               
              #jack1                                                                                             

0             1            2             3             4           5          6           7        8         9   
\end{BVerbatim}
\end{center}
\caption{\small The best alignment found by SP61 with patterns representing classes of entity Old and the pattern `Jack stethoscope black-bag fair-hair blue-eyes Dorking' in New. In the alignment, the symbol `beard' is shorthand for the ability to grow a beard.}
\label{alignment_1}
\end{figure*}

Figure \ref{alignment_1} shows the best alignment found by SP61 with a set of patterns in Old describing classes of entity and the pattern `Jack stethoscope black-bag fair-hair blue-eyes Dorking' in New. By convention, the New pattern is always shown in column 0, with Old patterns in the remaining columns. Apart from this constraint, the ordering of patterns across the columns is entirely arbitrary.

To avoid creating an alignment that would be too big to fit on the page, many details have been omitted. There is, for example, nothing in this alignment to show the structure of a person's body or legs (column 2) or the nature of that person's nose or mouth (column 7).

In the majority of cases, a given pattern contains one or more symbols that serve as a means of identifying the pattern and referencing it from one or more other patterns within the system. An example from Figure \ref{alignment_1} is the pair of symbols `person' and `\#person' at the beginning and end of the pattern shown in column 2. These have a r{\^o}le in the system which is similar to the way in which symbols such as `$<$person$>$' and `$<$/person$>$' would be used in XML. The XML style could be used instead. 

\subsection{Recognition and retrieval}

The most natural interpretation of an alignment like the one shown in Figure \ref{alignment_1} is that it is the result of a process of pattern {\em Recognition} and information {\em Retrieval} and, as discussed below, inference or {\em Reasoning} (RRR).

In this example, the New pattern may be seen as fragmentary information about some unknown entity and the alignment shows that it has been recognised as being one Jack Jones of Dorking (column 1). At the same time, the unknown entity is recognised as belonging to the class `person' (column 2), the class `male' (`gender' in column 8), and the class `doctor' (`profession' in column 3).

The alignment may also be seen as the result of a process of information retrieval, in the manner of `query-by-example' in a conventional database.

This process of recognition/retrieval is `fuzzy' in the sense that it can find good partial matches between patterns as well as exact matches. This capability stems from the use of an improved form of `dynamic programming' at the heart of the SP61 model.

\subsection{Reasoning}

Any symbol in an Old pattern in an alignment that is {\em not} matched with a symbol in New represents an {\em inference} that may be drawn from the alignment.

From the alignment in Figure \ref{alignment_1} we can infer {\em inter alia} that `Jack Jones' is (probably) a doctor with the usual attributes of doctors (shown as `...') and that he (probably) has a head, body and legs and other attributes of people. In effect, Jack Jones `inherits' the attributes of `doctor' and `person' in the manner of inheritance in object-oriented systems.

The alignment in Figure \ref{alignment_1} also illustrates another kind of `deduction'. The pattern shown in column 9 records the well-known association between the name `Jack' and the attribute `male'. This provides an inferential link between `Jack' in column 0 and the pattern representing the male gender shown in column 8. Notice also how the male characteristics, `beard' and `deep' (voice) which are part of that pattern are tied in to their proper places in the patterns for `head' (column 7) and `person' (column 2), respectively.

These are just two relatively simple examples of the way in which the ICMAUS framework supports reasoning. The system lends itself most naturally to probabilistic reasoning including inheritance of attributes and probabilistic `deduction' (as we have seen), and also probabilistic chains of reasoning, abductive reasoning, nonmonotonic reasoning and `explaining away'. Each pattern in Old has an associated frequency of occurrence in some domain and this information can be used by the SP61 model to calculate probabilities of alternative inferences drawn by the system. A relatively full account of how the system can perform these kinds of reasoning and how probabilities are calculated is presented in \cite{wolff_1999_prob} (see also \cite{wolff_2001_igpl}).

Although the system is fundamentally probabilistic, it can in principle also support classical `exact' modes of reasoning, as described in \cite{wolff_icmaus_overview}.

\subsection{Representation of knowledge}

Although all knowledge in the system is represented with `flat' patterns, the way in which the system builds multiple alignments means that these flat patterns can be used to represent a wide variety of other structures including context-free and context sensitive grammars (see \cite{wolff_2000}), class-inclusion hierarchies and part-whole hierarchies (discussed below), if-then rules, networks and trees \cite{wolff_1999_prob,wolff_icmaus_overview,wolff_2001_igpl}, and others. This subsection briefly reviews a selection of issues related to the representation of ontologies.

\subsubsection{Classes and instances}

In the ICMAUS framework, there is no formal distinction between a `class' and an `instance'---both these things are represented with patterns. In Figure \ref{alignment_1}, a pattern like the one for `person' (column 2) is most naturally interpreted as a class while the pattern representing `Jack Jones' (column 1) may be seen as an instance.

Philosophically, there is some justification for blurring the traditional distinction between `class' and `instance'. Although a given person like `Jack Jones' may seem like a singleton, any such concept is derived from a myriad of individual `perceptions': Jack going for a walk, Jack in the office, Jack eating a sandwich, and so on. There is no real justification for making a fundamental distinction between this kind of complex concept and other concepts that are more commonly regarded as classes.

\subsection{Properties and values}

The attributes of a class may be represented by literals such as `stethoscope' in column 3 of Figure \ref{alignment_1} or they may be represented by a pair of symbols such as `name \#name' in column 2, or `hcolour \#hcolour' in column 5. Any pair of symbols like that may function as a `slot' or `variable' or, in the terms of RDF, a `property'. And each such slot may receive a range of alternative `values', such as `Jack Jones' for the `name \#name' slot and `fair-hair' for the `hcolour \#hcolour' slot.

\subsubsection{Intensional and extensional representation of classes}

In the ICMAUS framework, classes may be represented intensionally with patterns like the ones for `person' or `profession' and they may also be represented extensionally. An example of the latter in Figure \ref{alignment_1} is `hcolour', already mentioned, where the range of possible values for that property is represented in Old by patterns such as `hcolour fair-hair \#hcolour', `hcolour black-hair \#hcolour' and `hcolour red-hair \#hcolour'. Any other class may be represented extensionally in a similar way.

\subsubsection{Cross-classification and multiple inheritance}

The system accommodates a strict hierarchy of classes and subclasses but it also supports cross-classification at any level in a hierarchy, with `multiple' inheritance from overlapping classes. In Figure \ref{alignment_1}, the classes `doctor' (profession) and `male' (gender) are both subclasses of `person' but neither one is a subclass of the other. These are classes that overlap in the world as we know it and any given person may inherit attributes from either one, or both, or neither.

\subsubsection{Properties, values and classes}

In our everyday `natural' concepts, it is often difficult to make a firm distinction between `properties', `values' and `classes'. A concept like `doctor' may be regarded as a property of some individual person (with a range of alternative values such as `paediatrician', `haematologist' etc) or, as in Figure \ref{alignment_1}, it may be seen as a value of a property such as `profession', and in all cases it may be regarded as a class.

The ICMAUS system reflects this aspect of `properties', `values' and `classes' by making no formal distinction between them.

\subsubsection{Class-inclusion relations, part-whole relations and their integration}

Apart from the class-inclusion relations that can be seen in Figure \ref{alignment_1}, the alignment shows how a person is composed of a `head', `body' and `legs', how the `head' is composed of `hair', `eyes', `nose' etc, with further detail for `hair' and `eyes'. A more comprehensive example would expand the hierarchical structure of other parts such as `body' and `legs'.

As discussed below, the ICMAUS framework allows class-inclusion relations to be integrated with part-whole relations in a way which is, arguably, more satisfactory that with RDF+ and some other languages in the object-oriented tradition.

\subsubsection{The order of symbols in patterns and alignments}

In the building of multiple alignments, the order of symbols in patterns is important. This is appropriate for many applications (e.g., natural language processing) but in other applications (e.g., medical diagnosis) it is useful to be able to treat patterns as unordered collections of symbols. This can be achieved by the provision of a `framework' pattern as explained in \cite{wolff_1999_prob}.

\subsubsection{Syntax and semantics}

In the ICMAUS framework, there is no formal distinction between syntax and semantics. In some applications (e.g., natural language processing) any given symbol or pattern may be assigned informally to one r{\^o}le or the other---but any such distinction relates to the particular application rather than the system itself.

Any symbol or combination of symbols may be seen to represent the `meaning' of one or more other symbols with which it is associated. In this connection, r{\^o}les can easily be reversed: `smoke' normally means `fire' but (less certainly) `fire' can also mean `smoke'.

\section{RDF+ and ICMAUS compared}

In this discussion, RDF+ will be equated with RDF, RDFS and DAML+OIL but similar remarks probably apply to other RDF-based proposals. RDF provides a syntactic framework, RDFS adds modelling primitives such as `instance-of' and `subclass-of', and DAML+OIL provides a formal semantics and reasoning capabilities based on Description Logics (DLs), themselves based on predicate logic.

\subsection{Representation of knowledge}

The ICMAUS theory incorporates a theory of knowledge in which syntax and semantics are integrated and many of the concepts associated with ontologies are implicit. There is no need for the explicit provision of constructs such as `Class', `subClassOf', `Type', `hasPart', `Property', `subPropertyOf', or `hasValue'. Classes of entity and specific instances, their component parts and properties, and values for properties, can be represented in a very direct, transparent and intuitive way, as described above. In short, the ICMAUS approach allows ontologies to be represented in a manner that appears to be simpler and more comprehensible than RDF+, both for people writing ontologies and for people reading them.

\subsubsection{Class-inclusion relations and part-whole relations}

There is a part-whole relation between the intensional description of a class and the class itself. In the ICMAUS framework, this part-whole relation can be exploited to represent part-whole hierarchies that are integrated with class-inclusion hierarchies and participate fully in inheritance.

RDF+ represents part-whole relations using constructs such as `hasPart' or `partOf'. In other object-oriented languages (e.g., Simula and C++) this kind of mechanism can mean restrictions on the inheritance of information from part-whole hierarchies. It is not entirely clear whether or not these restrictions also apply to RDF+.

\subsection{Reasoning}

Reasoning in RDF+ is based on predicate logic and belongs in the classical tradition of monotonic deductive reasoning in which propositions are either {\em true} or {\em false}. By contrast, the main strengths of the ICMAUS framework are in probabilistic `deduction', abduction and nonmonotonic reasoning. There is certainly a place for classical styles of reasoning in the Semantic Web but probabilistic kinds of reasoning are needed too. The following subsections expand on these points.

\subsubsection{Probabilistic and exact reasoning}

At the heart of the ICMAUS system is an improved version of `dynamic programming' that can find good partial matches between patterns as well as exact matches. The building of multiple alignments involves a search for a global best match amongst patterns and accommodates errors of omission, commission and substitution in New information or Old information. This kind of capability---and the ability to calculate probabilities of inferences---seems to be outside the scope of `standard' reasoning systems currently developed for RDF+ (see, for example, \cite{horrocks_2002}). It has more affinity with `non-standard inferences' for DLs (see, for example, \cite{baarder_etal_1999,brandt_turhan_2001}).

Given the anarchical nature of the Web, there is a clear need for the kind of flexibility provided in the ICMAUS framework. It may be possible to construct ontologies in a relatively disciplined way but ordinary users of the web are much less controllable. The system needs to be able to respond to queries in the same flexible and `forgiving' manner as current search engines. Although current search engines are flexible, their responses are not very `intelligent'. The ICMAUS framework may help to plug this gap.

\subsubsection{Deduction and abduction}

With the Semantic Web, we may, on occasion, wish to reason deductively (e.g., a flat battery in the car means that it will not start) but there will also be occasions when we wish to reason abductively (e.g., possible reasons for the car not starting are a flat battery, no fuel, dirty plugs etc). For each alternative identified by abductive reasoning, we need some kind of measure of probability.

Although classical systems can be made to work in an abductive style, an ability to calculate probabilities or comparable measures needs to be `bolted on'. By contrast, the ICMAUS system accommodates `backwards' styles of reasoning just as easily as `forwards' kinds of reasoning and the calculation of probabilities derives naturally from the MLE principles on which the system is founded.

\subsubsection{Monotonic reasoning, nonmonotonic reasoning and default values}

If we know that all birds can fly and that Tweety is a bird then, in classical logic, we can deduce that Tweety can fly. The `monotonic' nature of classical logic means that this conclusion cannot be modified, even if we are informed that Tweety is a penguin.

In the Semantic Web, we need to be able to express the idea that {\em most} birds fly (with the default assumption that birds can fly) and we need to be able to deduce that, if Tweety is a bird, then it is {\em probable} that Tweety can fly. If we learn subsequently that Tweety is a penguin, we should be able to revise our initial conclusion.

This kind of nonmonotonic reasoning is outside the scope of classical logic but is handled quite naturally by the ICMAUS framework (see \cite{wolff_1999_prob,wolff_2001_igpl}).

\section{Conclusion}

One of the strengths of the ICMAUS framework is that it allows ontological knowledge to be represented in a manner that is relatively simple and comprehensible. It has an ability to cope with uncertainties in knowledge and it can perform the kinds of probabilistic reasoning that seem to be needed in the Semantic Web.

The ICMAUS framework also has strengths in other areas---such as natural language processing, planning and problem solving, and unsupervised learning---that may prove useful in the future development of the Semantic Web.

\section*{Acknowledgements}

I am very grateful to Manos Batsis, Pat Hayes, Steve Robertshaw and Heiner Stuckenschmidt for 
constructive comments that I have received on these ideas.


\begin{thebibliography}{10}

\bibitem{baarder_etal_1999}
F.~Baader, R.~K{\"u}sters, A.~Borgida, and D.~L. Mc{G}uinness.
\newblock Matching in description logics.
\newblock {\em Journal of Logic and Computation}, 9(3):411--447, 1999.

\bibitem{brandt_turhan_2001}
S.~Brandt and A.-Y. Turhan.
\newblock Using non-standard inferences in description logics: what does it buy
  me?
\newblock In {\em Proceedings of {KI}-2001 Workshop on Applications of
  Description Logic ({KIDLWS}'01)}, 2001.

\bibitem{broekstra_etal_2002}
J.~Broekstra, M.~Klein, S.~Decker, D.~Fensel, F.~{van Harmelin}, and
  I.~Horrocks.
\newblock Enabling knowledge representation on the web by extending {RDF}
  schema.
\newblock {\em Computer Networks}, 39:609--634, 2002.

\bibitem{fensel_etal_2001}
D.~Fensel, I.~Horrocks, F.~{van Harmelin}, D.~L. Mc{G}uinness, and P.~F.
  {Patel-{S}chneider}.
\newblock {OIL}: an ontology infrastructure for the {S}emantic {W}eb.
\newblock {\em IEEE Intelligent Systems}, 16(2):38--45, 2001.

\bibitem{horrocks_2002}
I.~Horrocks.
\newblock Reasoning with expressive description logics: theory and practice.
\newblock In A.~Voronkov, editor, {\em Proceedings of the 18th International
  Conference on Automated Deduction (CADE-18)}, volume 2392 of {\em Lecture
  Notes in Artificial Intelligence}, pages 1--15. Springer-Verlag, 2002.

\bibitem{li_vitanyi_1997}
M.~Li and P.~Vit\'{a}nyi.
\newblock {\em An Introduction to Kolmogorov Complexity and Its Applications}.
\newblock Springer-Verlag, New York, 1997.

\bibitem{mcguinness_2002}
D.~L. McGuinness, R.~Fikes, J.~Hendler, and L.~A. Stein.
\newblock {DAML+OIL}: an ontology language for the {S}emantic {W}eb.
\newblock {\em IEEE Intelligent Systems}, 17(5):72--80, 2002.

\bibitem{rissanen_1978}
J.~Rissanen.
\newblock Modelling by the shortest data description.
\newblock {\em Automatica-J, {IFAC}}, 14:465--471, 1978.

\bibitem{solomonoff_1964}
R.~J. Solomonoff.
\newblock A formal theory of inductive inference. parts {I} and {II}.
\newblock {\em Information and Control}, 7:1--22 and 224--254, 1964.

\bibitem{wallace_boulton_1968}
C.~S. Wallace and D.~M. Boulton.
\newblock An information measure for classification.
\newblock {\em Computer Journal}, 11(2):185--195, 1968.

\bibitem{wolff_1999_comp}
J.~G. Wolff.
\newblock {`Computing'} as information compression by multiple alignment,
  unification and search.
\newblock {\em Journal of Universal Computer Science}, 5(11):777--815, 1999.
\newblock Copy: www.jucs.org/jucs\_5\_11.

\bibitem{wolff_1999_prob}
J.~G. Wolff.
\newblock Probabilistic reasoning as information compression by multiple
  alignment, unification and search: an introduction and overview.
\newblock {\em Journal of Universal Computer Science}, 5(7):418--462, 1999.
\newblock Copy: www.jucs.org/jucs\_5\_7. The three articles on which this
  article is based may be obtained from www.iicm.edu/wolff/1998a, b, c.

\bibitem{wolff_2000}
J.~G. Wolff.
\newblock Syntax, parsing and production of natural language in a framework of
  information compression by multiple alignment, unification and search.
\newblock {\em Journal of Universal Computer Science}, 6(8):781--829, 2000.
\newblock Copy: www.jucs.org/jucs\_6\_8. Three articles that are the basis of
  this article may be obtained from www.iicm.edu/wolff/1998d1, d2, d3.

\bibitem{wolff_2001_igpl}
J.~G. Wolff.
\newblock Information compression by multiple alignment, unification and search
  as a framework for human-like reasoning.
\newblock {\em Logic Journal of the IGPL}, 9(1):205--222, 2001.
\newblock First published in the {\it Proceedings of the International
  Conference on Formal and Applied Practical Reasoning (FAPR 2000)}, September
  2000, ISSN 1469--4166. Copy: www.cognitionresearch.org.uk/papers/pr/pr.htm.

\bibitem{wolff_icmaus_overview}
J.~G. Wolff.
\newblock Information compression by multiple alignment, unification and search
  as a unifying principle in computing and cognition.
\newblock {\em Artificial Intelligence Review}, 19(3):193--230, 2003.
\newblock Copy: www.cognitionresearch.org.uk/papers/overview/overview.htm.

\end{thebibliography}
\end{document}